\documentclass{sig-alternate}

\usepackage[latin1]{inputenc}

\begin{document}
%
\conferenceinfo{GECCO'06,} {July 8--12, 2006, Seattle, Washington, USA.}
\CopyrightYear{2006}
\crdata{1-59593-186-4/06/0007}

\title{On the Benefits of Inoculation, an Example in Train Scheduling}

%
%

\numberofauthors{2}
%


%

\author{
\alignauthor Yann Semet\\
       \affaddr{INRIA Futurs}\\
       \affaddr{LRI - Bât. 490 - Université Paris-Sud}\\
       \affaddr{Orsay, FRANCE}\\
       \email{semet@lri.fr}
\alignauthor Marc Schoenauer\\
       \affaddr{INRIA Futurs}\\
       \affaddr{LRI - Bât. 490 - Université Paris-Sud}\\
       \affaddr{Orsay, FRANCE}\\
       \email{marc@lri.fr}
}

\date{}

\maketitle

\begin{abstract}

The local reconstruction of a railway schedule following a small perturbation of the traffic, seeking minimization of the total accumulated delay, is a very difficult and tightly constrained combinatorial problem. Notoriously enough, the railway company's public image degrades proportionally to the amount of daily delays, and the same goes for its profit!

This paper describes an inoculation procedure which greatly enhances an evolutionary algorithm for train re-scheduling. The procedure consists in building the initial population around a pre-computed solution based on problem-related information available beforehand.

The optimization is performed by adapting times of departure and arrival, as well as allocation of tracks, for each train at each station. This is achieved by a permutation-based evolutionary algorithm that relies on a semi-greedy heuristic scheduler to gradually reconstruct the schedule by inserting trains one after another.

Experimental results are presented on various instances of a large real-world case involving around 500 trains and more than 1 million constraints. In terms of competition with commercial mathematical programming tool ILOG CPLEX, it appears that within a large class of instances, excluding trivial instances as well as too difficult ones, and with very few exceptions, a clever initialization turns an encouraging failure into a clear-cut success auguring of substantial financial savings.

\end{abstract}

\category{I.2.8}{Computing Methodologies}{Artificial Intelligence: Problem Solving, Control Methods, and Search}

\terms{Algorithms}


\keywords{Inoculation, Scheduling, Trains, Permutations}


\section{Introduction}

\subsection{Context}

Punctuality stands out as an important
factor of consumer satisfaction and plays a key role in the railway
world race for a competitive edge. Moreover, any delay for a given
train results in additional occupation of resources, and hence in
financial losses. The problem dealt with 
here was provided, together with real world datasets, by SNCF, the
French national railroad company. It is concerned with the diminution
of resulting delays in case of small 
perturbations of the traffic: each minute of delay costs SNCF
about 1000 Euros \cite{RailEtRecherche}! More precisely, when for some
operational reason, a train is delayed for a few minutes at some
point of the network (unavailable track, malfunctioning equipment,
etc.), it involves taking, within a few minutes, the right decisions for space-wise and
time-wise neighbouring trains so that the consequence is minimal in
terms of accumulated delays, while enforcing a great deal 
of operational, commercial and safety constraints. The problem at hand is therefore a difficult,
combinatorial, constrained optimisation problem, equivalent to some form of job-shop scheduling problem.

At the moment, this problem is
solved in real time by human operators who make use of
expert knowledge and common sense to perform optimisation. A first
scientific investigation of the problem by SNCF has led to its description and
solving as a Mixed Integer Program (MIP) with commercial tool 
ILOG CPLEX. 
Obtained results are satisfactory for they show the adequacy of
the chosen model and prove its computational tractability but do not
meet criteria for real-time exploitation and suggest weakness with
respect to combinatorial explosion. They indeed show that several
hours are necessary on a fast computer to reach acceptable solutions
for average-size instances of the problem. Additionally, computational
efficiency seems to be greatly affected by growth of the instance in
either size or complexity, which eventually leads to
intractability. 

A second investigation, source of the present work, was then launched at SNCF's initiative. It lead to the problem's formulation and solving in evolutionary terms based on an indirect approach and a permutation representation, both described below. Early results are reported in \cite{semet:CEC05} and show very encouraging results on a single, large instance: the evolutionary algorithm is much faster than CPLEX early in the search but systematically converges at suboptimal levels of fitness. That work also showed how the evolutionary algorithm could be fruitfully hybridized with CPLEX to yield an efficient memetic algorithm.

The present paper details results obtained on many different instances and investigates a new, surprisingly efficient non-random initialization procedure for the evolutionary algorithm, which makes it outperform CPLEX on most instances.

\subsection{State of the Art}

\subsubsection{On Competent Initialization}

Most EC practitioners working on real-world applications will agree that ``good stuff'' will happen only when the algorithm is properly fed with relevant domain knowledge and that pure ``black-box'' optimization is bound to lead nowhere in most cases. There are many ways to achieve proper domain knowledge supply, among which stand out two for their importance and wideness of use: coming up with a concise and informative representation (notably providing tight linkage) and designing intelligent operators which perform competent recombination or efficient mutative exploration.

Another, often neglected way to provide problem-specific knowledge is to modify the algorithm's initialization phase heuristically. Common ways to do this include biasing random sampling toward more relevant zones (by informatively picking adequate parameter ranges for instance) and seeking to provide a richer ``primordial soup'' of good building blocks (see e.g., Messy GAs \cite{Kargupta:PhD}). 

Few works exist on the specific subject of improving Initialization while we argue it is one of the most efficient and easy way to improve an evolutionary algorithm's efficiency. A notable exception is the work by Surry \cite{SurryPhD,Surry:inoculation96} which provides both an overview of real-world application examples and an in-depth discussion of the issues involved, like premature convergence and population diversity.

When it happens to be possible, an excellent way to initialize well is to build the initial population around a previously known good solution. This is the approach we are taking although both the purpose and the technicalities slightly differ from the norm. Instead of using an existing good solution to the actual problem or building one using constructive heuristics or other search methods, we actually pre-solve part of the problem once and for all using our evolutionary algorithm. Our strategy is indeed a {\bf Divide\&Conquer} one: as described in section \ref{sec:init}, the problem can be decomposed in three parts, two of which can easily be solved beforehand, which results in an excellent starting point for the actual optimization phase.

\subsubsection{On Evolutionary Train Scheduling}

A number of teams worked on the particular case of train scheduling. We give a list, by no means exhaustive, of important examples. Caprara et al. \cite{caprara01solution} offered a MIP formulation of the problem along with a solving algorithm which makes use of graph-theoretic techniques and Lagrangian relaxation. Parkes and Ungar \cite{ParkesTrain} use market algorithms where trains are represented by virtual agents that competitively interact for the allocation of resources attributed by an efficient auction-based system. Kwan et al. propose a Co-evolutionary approach \cite{kwan:CEC03} for initial timetable generation. Finally, Juill\'e \cite{juille04trains} mixes permutation-based evolutionary search and constraint oriented programming to solve a bi-objective instance of the train timetabling problem in a decision support context. 
Following Juill\'e, our approach is an indirect one: the genotype is an
ordering of the trains (a permutation), and all constraints
are handled in some scheduler, i.e. during the morphogenesis process
transforming the permutation (i.e. the genotype) into a valid schedule.

\section{The Problem}
\label{sec:problem}

\subsection{Degrees of Freedom}

\begin{figure}[htbp]
    \begin{center}
        \epsfig{file=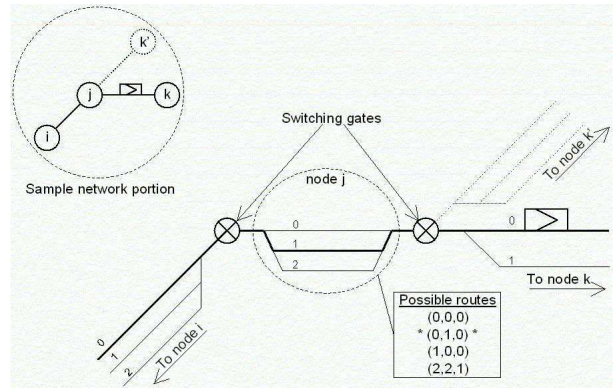,width=8cm}      
      \caption{A sample network portion - A train, coming from node $i$, is on its way to node $k$ after having gone through node $j$. It chooses the second possible route or set of incoming/inside/outgoing tracks: $(0,1,0)$.}
      \label{fig:samplePortion}
    \end{center}
\vspace{-0.4cm}
\end{figure}

As illustrated by Figure \ref{fig:samplePortion}, the railway network can be seen as a graph where nodes are stations or
switchings and where interconnecting edges eventually hold several
tracks for trains to use. Each train $c$ ($C$ being the set of all
trains\footnote{We choose $c$ instead of $t$ to avoid confusion with
time.}) has a fixed ordered list of nodes to visit $I(c)$. 
There
are three degrees of freedom for each train at each at node: times of
arrival $a$, departure $d$ and track choice $r$ (for
\emph{route}).

$a$ and $d$ are integers giving the number of seconds
elapsed since an arbitrary temporal origin.

Track choice actually
implies three tracks to be chosen: one in both incoming and outgoing
edges (resp. $u_{inc}$ and $u_{out}$) and one inside the node
($u$). These three decisions are linked by underlying physical
constraints: picking, for instance, a particular incoming track
restricts the number of possible subsequent tracks inside the node and
in the outgoing edge. Each node therefore holds a list of triplets
$(i,j,k)$ indicating the physically possible combinations among which a
choice has to be made.

To sum up, a schedule is completely defined by assigning values to all degrees of
freedom, i.e., for each train $c$ at each node $i$: $a(c,i) \in \mathbb{N}$, $d(c,i) \in \mathbb{N}$ and $r(c,i)=(u_{inc},u,u_{out}) \in \mathbb{N}^3$.



The problem at hand is to reconstruct a valid schedule after some
incident. The initial schedule that all trains had been assigned
before the incident  will be referred to by zero-indices
(e.g. $a_0(c,i), \ldots$).

\subsection{Visualisation}

\begin{figure}[htbp]
    \begin{center}
        \epsfig{file=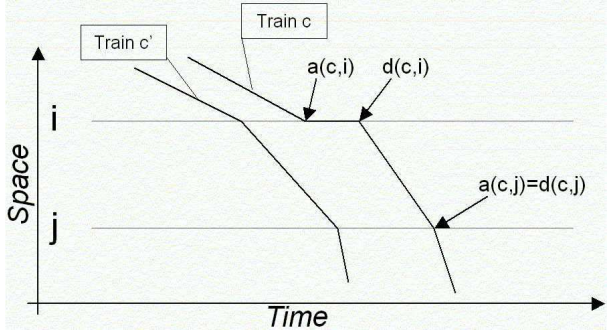,width=8cm}      
      \caption{A space/time diagram - Two trains are represented, both going from node $i$ to node $j$. Train $c$ stops at node $i$ whereas $c'$ doesn't.}
      \label{fig:spaceTime}
    \end{center}
\vspace{-0.4cm}
\end{figure}

As illustrated by self-explanatory Figure \ref{fig:spaceTime},
schedules can be represented using space/time diagrams. This visual
representation is very useful to railway scheduling experts to get a
global glimpse of a complex network and to quickly detect
abnormalities. The reader might also find it useful to think the
equations given below in those visual terms and to see it as a visual
representation of the phenotype. 

\subsection{Constraints}

This sections gives a brief enumeration of the constraints limiting the aforementioned degrees of freedom. Please refer to \cite{semet:CEC05} for an exhaustive and more detailed list. All constraints are hard constraints and cannot possibly be violated, mostly for safety reasons. This set of constraints is not exactly exhaustive with respect to real world operational conditions but gets pretty close, which definitely places the problem outside the ``toy problem'' category.

For clarity, quantifiers (e.g. $\forall c$, $\forall i$, etc.) are omitted in all equations given below, assuming inequalities apply to the obvious relevant sets. For instance, safety spacing constraints only apply between trains that use the same resource (node or track section) at some point. All Greek letters, introduced in alphabetical order, will represent prescribed constants for spacing intervals.

\begin{enumerate}
\item {\bf Initial Times.} These constraints are the most obvious ones: trains cannot arrive or leave a node earlier than specified in the initial schedule ($a(c,i)\ge a_0(c,i)$, $d(c,i)\ge d_0(c,i)$).

\item {\bf Stopping Time.} For both maintenance and commercial reasons, trains stopping times are both upper and lower bounded  ($\alpha_{min}(c,i) \leq d(c,i)-a(c,i)\leq \alpha_{max}(c,i)$).

\item {\bf Speed.} According to physical contingencies such as engine power, trains need a certain time to cover the distance between two nodes $i$ and $i'$ ($a(c,i')-d(c,i) \geq \beta(c,i\to i')$).

\item {\bf Safety Spacing.} At all times, there must be sufficient distance between any pair of trains so that either has enough time to undertake and complete emergency braking procedures. This applies for edges as well as for nodes (e.g. for nodes : $a(c',i)\geq d(c,i)+\gamma(c,c',i)$).

\item {\bf Connections} Connecting trains must be coordinated and give each other enough time at connecting nodes to ensure proper transfer of passengers.

\item {\bf Switching Gates} Two edges connected to one same node can share switching gates. These gates have limited capacities, being able, for example, to handle only one train at a time. Spacing constraints appear as a consequence to ensure proper use of this shared physical resource. These constraints are responsible for a very large part of the problem's combinatorial complexity.

\end{enumerate}

\subsection{Objective Function}

The goal of the optimisation procedure is to minimise the total accumulated delay, i.e. for all trains at all nodes, the difference between the actual time of arrival and the theoretical one. One should note that this is only used as a first step and will probably be improved in future studies. Many refinements are indeed obviously desirable like having trains of smaller importance matter less or seeking well spread rather than strictly minimal delay (a few slightly delayed trains would be preferable to one strongly delayed).

Using the above notations, and considering the fact that the theoretical schedule is fixed, the fitness function can be written as
\begin{eqnarray}
        f=\sum_{c \in C}{\sum_{i \in I(c)}{a(c,i)}}
\end{eqnarray}

\section{The Permutation Approach}
\label{sec:approach}

We use an indirect approach based on a permutation representation. This approach is classical for Job-Shop or Time\-tabling problems. A mapping function, generally called a {\em scheduler}, constructs the solution by iterative greedy insertion of elements (tasks, planes, \ldots in our case trains) in the order dictated by the permutation. \emph{Greedy} means that when a train is placed in the schedule, it uses the available resources, as left by previously scheduled trains, in the optimal way. The underlying idea is that the evolutionary optimisation process should come up with the permutation representing the order in which trains should be allowed to use available resources (tracks, gates, \ldots) and the priority according to which they could be forced to wait for fluidification purposes. The order resulting from the Evolutionary Algorithm should form the basis of an efficient response to the perturbation that created the problem.

The evolutionary algorithm thus evolves a population of
permutations. These permutations (the genotypes of the Evolutionary
Algorithm) are turned into proper schedules (the phenotypes) by the
scheduler. Such schedules are then 
evaluated for total accumulated delay, with respect to the initial,
unperturbed schedules, to provide fitness. This population goes
through the traditional evolutionary loop (see Section
\ref{sec:evo}) until it 
reaches a satisfactory solution. 

\subsection{A Semi-Greedy Scheduler}
The ``scheduler'' procedure is at the very heart of our algorithm and forms the basis on which much of the resulting performance is going to come from.

The role of the scheduler is to read the permutation of trains and to place them, each in turn, in the schedule while respecting all of the specified constraints. The following sections explain how constraints are checked for and how conflicts are detected and eliminated. 

~
\subsubsection{Main Loop}

A train is inserted node by node. For each node, a loop goes through all the possible routes (i.e. triplets of incoming, inside and outgoing tracks). The route allowing for the earliest departure time is chosen and one moves on to the next node.

For each route, the arrival and departure times are initialised to their original, unperturbed value or, at departure time set for the previous node plus the minimum time it takes to get to the current node. The constraint checking loop then begins: for each kind of constraint in turn, one checks for violation. If it happens, what we call a \emph{conflict} occurs. This conflict is solved using either method described below, a flag is raised and the loop is started again.  The loop naturally goes on until no violation flag is raised.

\subsubsection{Solving Conflicts}

When a constraint is violated by a train being scheduled (noted below as train $A$), the conflict (usually with a previously scheduled train, noted below as train $B$) can be solved in two ways. If it is possible, $A$ moves its $a$ and $d$ variables forward in time until the violation threshold is overcome. If such a move is impossible (for example because it creates another fatal conflict), $B$ is removed from the schedule, thus solving the conflict, and placed on top of the ``scheduling stack'' to be re-inserted right after one is done with $A$. The latter way is called the ``kick'' procedure and can be called for only a limited number of times per train in order to avoid infinite unscheduling cycles.

\subsection{Issues}

Three difficulties are created by using this indirect approach. The first one is that removing previously scheduled trains destructurates the permutation because the final scheduling order is not the one initially present in the permutation. This makes the optimisation process harder to understand and therefore to enhance. The second difficulty lies in the fact that not all of the permutations are feasible individuals: for some of them, a few trains cannot enter the schedule without violating constraints ; these individuals have a strongly penalized fitness and disappear early in evolution. Finally, the combination of an indirect approach and a fast semi-greedy scheduler (i.e. optimal at node level, locally, as opposed to a much slower actually greedy scheduler which would be optimal globally, at train level) only leads to suboptimal solutions which is not a major problem as the EA goal is to find excellent, not optimal, solutions very quickly. These issues are discussed extensively in \cite{semet:CEC05}.

\section{The Evolutionary Algorithm}
\label{sec:evo}

\subsection{Representation}

The problem for the Evolutionary Algorithm is to find an optimal
permutation such that it results in an
optimal schedule when fed into the scheduler.
We chose to use a direct representation: permutations are
straightforwardly encoded using a sequence of integers. 
Other representations were considered, like the Edge Representation
\cite{radcliffeBinMut}, 
or the Random Keys \cite{RKGA}.
However,  edges are here meaningless, and Random Keys would add
yet another mapping step between genotype and phenotype,
making it even harder to understand how the EA navigates through
the search space. Additionally, 
beyond its simplicity in terms of
implementation, direct representation is here better
suited for the design of specific, intelligent operators based on
expert knowledge of the problem, which is often the key to success
for difficult real-world problems.  

\subsection{Population and Selection Scheme}

A standard $(\mu+\lambda)$ replacement scheme is borrowed from Evolution Strategies: a population of $\mu$ parents produces $\lambda$ children using variation operators (see below). Among the set of both parents and offspring, the best $\mu$ individuals are chosen as the  new parents for the next generation.

A smoother replacement is the so-called \emph{Evolutionary Programming Tournament (EPT)}: each individual virtually encounters $S$ randomly chosen opponents, and scores 1 when it is fitter than them. The $\mu$ highest scores are selected to become the parents of the next generation. EPT tournament turned out to be essential with layer-based initialization schemes (see below) to allow low fitness random individuals to survive the initial competition with high fitness inoculated individuals.

Finally, we use {\emph{Ordered selection}}, which means that all members of the population are successively called to produce $\lambda$ children in turn. This allows for a diversity preserving low selection pressure -- the only selection pressure comes from the replacement procedure.

\subsection{Mutation Operator}
\label{sec:mutation}
Several traditional permutation-based variation operators have been implemented, but extensive experimentation showed that using a simple, although enhanced, swap mutation (see below for details)  was the best choice. Among other variation operators that have been tested are some variant of the Partially Matched Crossover (PMX \cite{Goldberg89}), the Uniform Order Crossover (UOX \cite{Goldberg89}), the Half-Swap mutation and a variety of enhancements inspired by metaheuristics (Tabu Search, Simulated Annealing, etc.). For space reasons, only the retained Swap Mutation will be detailed here.

This mutation simply consists in swapping two elements of the permutation, which means two trains exchanging their ranks in the permutation. To prevent it from being too disruptive, this swapping takes place within a {\bf restricted radius} $R$, which means the two swapped elements cannot be distant of more than $R$ permutation spots. This swapping operation can be repeated $T$ times.

Drawing an analogy from {\bf Simulated Annealing}, we set the trade-off between exploration and exploitation by controlling the number of times this swap is performed by each mutation operation. This parameter $T$ (for \emph{Temperature}) remains fixed during the first $n_0$ generations, then decreases according to the following monotonous function, which is calibrated by a 4-uple $(n_0,T_0,T_\infty,\gamma)$:


\begin{equation}
T'(n)=T_\infty+2(T_0-T_\infty)(1-\frac{1}{1+e^{-\gamma(n-n_0)}})
\end{equation}

Additionally, and following Radcliffe and Surry \cite{radcliffeBinMut}, we make the choice of the number of transpositions less deterministic and narrow by drawing it as a realisation of a {\bf binomial law} with mean $T$:  $T$ is still the most likely value, but neighbouring values are possible as well, and far values are not impossible. This is interesting especially late in the search when $T(n)$ has converged to $T_\infty$: although the search will concentrate on very local mutations, exploratory mutations nevertheless occasionally take place. 

\section{Inoculation}
\label{sec:init}

\subsection{Rationale}


The main idea behind this new initialization heuristic comes from realizing that in this context, the evolutionary algorithm actually solves three problems at the same time :

\vspace{0.3cm}
{\bf $P_1$ : Moving from completely random permutations to reasonably natural ones.} Scheduling trains in a completely random order makes little sense. First of all, obvious heuristics dictate natural orderings for many subsets of trains : earlier trains should be scheduled first, chains of connecting trains should be scheduled in order, important trains should come first, etc. But more importantly, the theoretical schedule, designed by real-world scheduling experts, is a nearly optimal one and benefits from a long history of day by day optimization which is reflected in the relative priorities given to certain pairs or subsets of trains, for example to access a given high-speed track. Starting from a random population, our algorithm has to rediscover all of this structure along the way (while optimizing) and come up with a set of permutations that correspond to natural schedules situated reasonably close to the theoretical, unperturbed schedule. 

{\bf $P_2$ : Correcting initially present minor violations.} This is peculiar to the problem and instance data studied here and is typical of real-world problems. It happens that the theoretical schedule does not respect all of the constraints specified in the mathematical model (two trains can, for instance, be scheduled to leave a given station a few seconds too close to each other than allowed by security regulations). Besides its job of minimizing the consequences of a perturbation, our gradual reconstruction algorithm must therefore detect and correct those small but numerous inconsistencies which can be found everywhere in the schedule both space-wise and time-wise.

{\bf $P_3$ : Solving the actual problem.} This is the actual, real-time purpose of the algorithm: finding a permutation that produces a schedule which provides a satisfying answer to the local perturbation that just occurred while respecting all of the aforementioned constraints.
\vspace{0.3cm}

The key idea is that $P_1$ and $P_2$ can be solved beforehand because they do not depend on the particular incident instance being solved. One just needs to run the Evolutionary Algorithm once with an empty perturbation, which means solving $P_1$ and $P_2$ with a fictitious, meaningless $P_3$. This results in a solution (the best individual obtained after a large number of generations) that is excellent with respect to $P_1$ and $P_2$ and that will be used as an inoculant (see below for details), or good starting point whenever a new instance of $P_3$ needs to be solved. This allows the algorithm to {\bf get to the point} directly, and optimize efficiently without being slowed down by difficulties previously found upon and solved.

This kind of Initialization is referred to as {\bf Inoculation} in the literature as introduced in \cite{SurryPhD,Surry:inoculation96}. The pre-calculated solution around which the initial population is built is called the {\bf inoculant}, noted $I_0$ below. The next section describes the various ways in which it can be done.

One should note that CPLEX also uses a constructive procedure to come up with a good intial solution situated close to the unperturbed schedule. Unfortunately, because of the indirect approach we take, this initial solution cannot be used as a starting point for the Evolutionary Algorihtm because we do not know how to turn it into a corresponding permutation.

\subsection{Variants on How to Inoculate}

We build the initial population used by the Evolutionary Algorithm by cloning and perturbing the inoculant as many times as needed. The perturbation, similarly to what the mutation does, is achieved by performing a chosen number $pR$ (for \emph{perturbation Radius}) of transpositions (swapping a randomly chosen pair of elements of the permutation). An inoculant $I_0$ perturbed $pR$ times is noted $I_0+pR$ below. The different initialization schemes are the following:


\vspace{0.3cm}
{\bf Mass Mutation $MM(pR)$}: The population is uniformly initialized with clones of $I_0$, each perturbed by $pR$ transpositions ($I_0+pR$). This is the most straightforward way to inoculate. We use $MM(3)$ by default.

{\bf Gradual Perturbation $GPer(pR_0,pR_{inc})$ }: As the initial population is being built, each individual is perturbed with an increasing radius. Individual number $k$ is thus $I_0+pR_0+(k-1).pR_{inc}$. The idea is to benefit from a large variety of possibilities ranging from close to far away from $I_0$ in order to get quick excellence while avoiding premature convergence. We use $GPer(0,1)$ by default.

{\bf Layer Initialization $LI(x_1/pR_1, \ldots, x_M/pR_M)$} Still in order to benefit from both the excellence of the zone close to $I_0$ and the diversity provided by randomness, we radicalise the previous method by cutting the initial population into layers of size $x_k$ (given in \%), each of them being uniformly perturbed with a radius of $pR_k$ transpositions. We actually used two schemes in turn:
	\begin{enumerate}
	\item {\bf Three Layers $T(33/0,33/10,33/500)$} The first layer contains unperturbed clones of the inoculant, the second layer, mildly perturbed ones, and the last layer, completely random permutations. 
	\item {\bf Two Layers $H(50/3,50/500)$} This scheme goes one step further toward radicalisation and gets rid of the intermediate layer in order to focus more clearly on the two sources of building blocks we are interested in: the inoculant and initial randomness. 
	\end{enumerate}
\section{Experimental Setup and Results}
\label{sec:results}

\subsection{Easy and Difficult instances}

Results reported here are obtained on a problem of size 541 (trains), on a large and complex railway network representing the suburban interconnections of a large city. It corresponds, in CPLEX terms, to 1 million variables and 3 million constraints.

Results vary greatly with respect to the perturbation instance considered. Influential factors are: the time at which the perturbation occurs, where it occurs and how long the corresponding train is delayed. We therefore compare results obtained on several such instances representing a variety of different cases. An illustrative example is instance A: it has medium/high difficulty and corresponds to a train being delayed for 10 minutes at 7am in a large node connecting suburban trains with intercity ones.

The duration of the delay (refereed to as $\delta$ below) is particularly important: as it grows, an instance can quickly go from very easy to very hard. Our algorithm behaves differently on easy and hard instances. An instance is classified as easy when CPLEX reaches the optimum almost immediately. To illustrate the difference, we compare results obtained on 4 pairs of instances, each of which has an easy version (B, C, D and E respectively) and a hard one obtained by increasing $\delta$ (resp. B',C',D' and E'). All results are discussed and considered but for the sake of clarity, only the most striking and illustrative ones are charted below.

\subsection{Experimental and parameter settings}

All results obtained by the Evolutionary Algorithm given below are averaged out of eleven runs. Dominance or equivalence of algorithmic variants are refereed by the Wilcoxon unsigned test with a 99\% confidence.

All tests are run on Pentium IV, 2GHz machines with 512MB RAM. 
On each instance, we independently run CPLEX for 24 hours and the Evolutionary Algorithm for 100 generations (around 2 hours). We report comparisons in terms of speed on the same machine over the first two hours of computation, which we observed as a sufficiently significant period of time.

We use a population of size $\mu=10$ parents, each producing 7 ($\lambda=70$) offspring. Extensive experimentation has led to these particular values but has also shown that the algorithm is very robust with respect to these parameters. For the present set of experiments, the mutation radius is not limited ($R=\infty$).

Table 1 details and gives convenient IDs to the various algorithmic variants described earlier and whose results are compared below. The ``T'' column describes the way in which the mutation strength $T$ is set during the run: $T$ is either constant or decreased as described in section \ref{sec:mutation} in which case a calibrating 4-uple is given; a $b$ subscript indicates the use of the binomial procedure described in Section \ref{sec:mutation}. The ``Init'' column states how the initial population is built around the previously obtained inoculant using parameters described above: perturbation radius for \emph{Mass Mutation}, proportions for \emph{Layers}, etc.

    \begin{table}[h]\label{tab:variants}
      \begin{center}
        \begin{tabular}{|c|c|c|}
	        \hline
	        \bf Variant ID & \bf T & \bf Init.\\
		\hline
	  	\hline 
	  	$MM$ & $4_b$ & 3 \\

		\hline 	
	  	$GPer$ & $4_b$ & (0,1) \\
		\hline
		$R$ & $(3,50,4,0.2)$ & 3 \\
\hline
$H$ & $4_b$ & (50/3,50/500) \\
\hline
$T$ & $4_b$ & (33/0,33/10,33/500)\\
\hline
$H+R$ & $(3,50,4,0.2)$ & (50/3,50/500) \\
\hline
$T+R$ & $(3,50,4,0.2)$ & (33/0,33/10,33/500)\\
          \hline
        \end{tabular}
      \end{center}
      \caption{Tested Algorithmic Variants}
    \end{table}

\subsection{Mass Mutation alone}

\begin{figure}[htbp]
    \begin{center}
        \epsfig{file=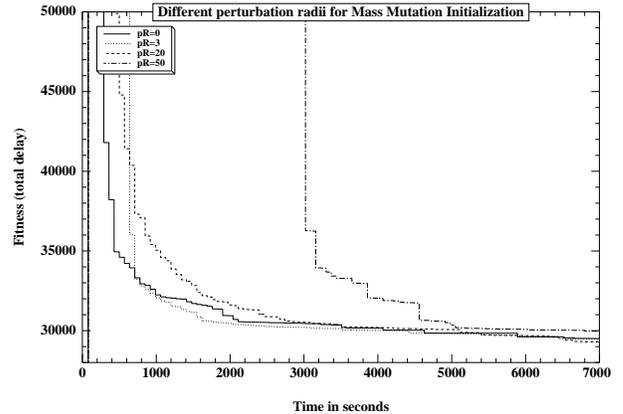,width=8cm}      
      \caption{Different perturbation values for Mass Mutation, all are approximately equivalent below 20, 50 illustrates a too strong perturbation momentarily kicking the EA outside of the feasible zone.}
      \label{fig:MMpR}
    \end{center}
\vspace{-0.4cm}
\end{figure}

We begin by the studying the most straightforward procedure: Mass Mutation ($MM$) and how different perturbation strengths ($pR$) can affect its efficiency. As can be seen in figure \ref{fig:MMpR}, $MM$ is very robust from that standpoint: performance is not affected by $pR$ when it lies below 20. Beyond a certain threshold however (between 20 and 50) the perturbation gets too strong and introduces too much randomness, placing the EA initially outside the feasible zone, which is reached back only after a little less than an hour.

\subsection{Alternative Schemes}

We proceed then to comparing the various inoculation strategies described earlier: Mass Mutation ($MM$), Gradual Perturbation ($GPer$), Two Layers ($H$) and Three Layers ($T$). Result vary across instances but the general trend is the following: 1) $MM$ is better than both $H$ and $T$, 2) $H$ and $T$ are alternatively better than one another and 3) $GPer$ is equivalent to $MM$.  
Because of this third result, confirmed with 99\% confidence on all available runs by the Wilcoxon test, we do not report $GPer$ results below. 

\subsection{Varying T}

This set of experiments seeks to establish whether the simulated annealing inspired decrease of the mutation strength $T$ described in Section \ref{sec:mutation} is useful as opposed to a constant value (usually 4) slightly randomized using a binomial law. We modify all algorithmic variants detailed above accordingly and compare them to $MM$.

\begin{figure}[htbp]
    \begin{center}
        \epsfig{file=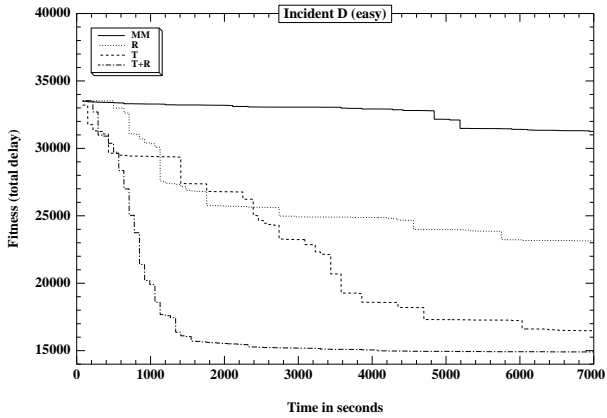,width=8cm}      
      \caption{Easy incidents:  algorithmic variants usually perform better than Mass Mutation alone. On this particularly striking example, a three layers scheme with simulated annealing inspired decrease of mutation strength does much better than all others.}
      \label{fig:recuit1}
    \end{center}
\vspace{-0.4cm}
\end{figure}

The results are different on easy instances and on hard ones. As exemplified by figure \ref{fig:recuit1}, algorithmic variants usually perform better than $MM$ on easy incidents. $T+R$ in particular outperforms or is equivalent to all other variants, including $MM$, on all easy incidents. Figure \ref{fig:recuit1} shows one particularly striking example where $T+R$ is better than both $T$ and $R$ alone and far better than $MM$. 

\begin{figure}[htbp]
    \begin{center}
        \epsfig{file=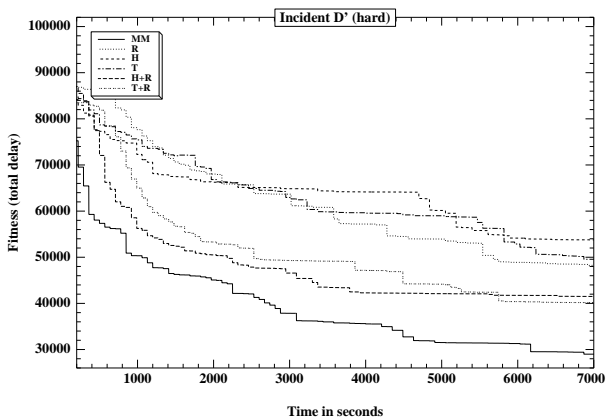,width=8cm}      
      \caption{On difficult incidents, Mass Mutation is usually the best choice, performing sometimes better, sometimes equivalently to other variants but never worse.}
      \label{fig:recuit2}
    \end{center}
\vspace{-0.4cm}
\end{figure}

On hard instances however, as shown in figure \ref{fig:recuit2}, the simple $MM$ with a small perturbation radius ($pR=3$), appears to be the best choice, as it performs significantly better than all others on several cases and not worse on the rest of the cases. $T+R$ and $H+R$ come second on this example, performing far better than $R$, $T$ or $H$ alone. This configuration is the the one observed most often.

Overall, unless one knows in advance what kind of incident has occurred, easy or hard, it is hard to tell which, of $T+R$ and $MM$ is the best choice. Difficult incidents being the Evolutionary Algorithm's target of choice, both intuition and simplicity suggest to keep $MM$ as the weapon of choice.

As to why such a dichotomy exist between easy and hard instances and to why particular variants are better suited than others in each case, one explanation seems to lie in the nature of the fitness landscape situated around the optimum. In easy instances, the feasible zone situated around the optimum is large and well structured. As such, it can benefit from variants such as $T+R$ which introduce interesting random blocks with a low risk of disruption or infeasibility and performs a wider sampling of the initial zone pointed out by the inoculant. Hard instances however have small feasible zones (many permutations lead to incomplete schedules) and a rougher landscape around their optima in the sense that small perturbations like a single short-radius transposition can make a good permutation infeasible. We argue that in such a case, a conservative, tightly focused \emph{Mass Mutation} scheme has therefore better chances of success than a more exploratory variant.

~

\subsection{Facing CPLEX}

The purpose of this work, conducted under an industrial contract, is to compare the performances of Evolutionary Algorithms to those of CPLEX on a large and difficult problem with the long term objective of real-time use. In this context, the initialization procedure we introduce here is of great help as it increases the algorithm's efficiency considerably.

Results obtained below compare CPLEX to the $MM$ variant of our algorithm. CPLEX is run for 24 hours but the comparison is made over the first two, the time it takes to the Evolutionary Algorithm to make a hundred generations and to converge to a seemingly global optimum. Nothing much happens after a 100 generations, i.e. after two hours, for the evolutionary algorithms whereas CPLEX can make (and often does) brutal leaps after several hours of absence of progress. As we are ultimately aiming for real-time performance, we chose to focus on the shortest possible significant period of time, two hours here, while keeping in mind what happens for CPLEX over the next 22 hours.

\begin{figure}[htbp]
    \begin{center}
        \epsfig{file=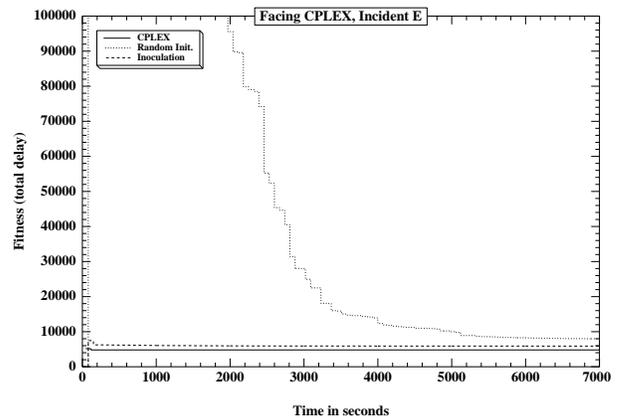,width=8cm}      
      \caption{On easy incidents, inoculation makes the EA competitive with CPLEX albeit not better.}
      \label{fig:facingCPLEX3}
    \end{center}
\vspace{-0.4cm}
\end{figure}

On easy instances to begin with, that is on instances where CPLEX, as can be seen on figure \ref{fig:facingCPLEX3} finds its best solution immediately or almost immediately, the Evolutionary Algorithm cannot hope to do any better than perfection and is, in its random initialization version, more than outperformed. Inoculation helps greatly reducing the gap, sometimes even almost suppressing it as illustrated in figure \ref{fig:facingCPLEX3}.

\begin{figure}[t]
    \begin{center}
        \epsfig{file=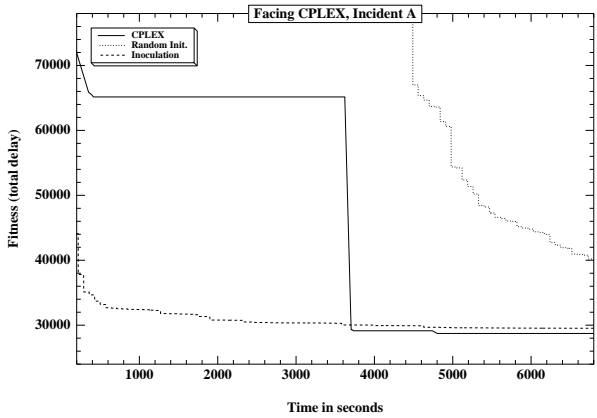,width=8cm}      
      \caption{The effect of Inoculation on a difficult instance: from an encouraging failure to a clear-cut success...}
      \label{fig:facingCPLEX1}
    \end{center}
\vspace{-0.4cm}
\end{figure}

On hard instances, the picture is entirely different. The example shown in figure \ref{fig:facingCPLEX1} is typical : the random initialization version is completely outperformed by CPLEX, only getting close in terms of fitness for a few minutes around the beginning of the second hour. The inoculated version however clearly outperforms CPLEX from the very beginning reaching almost optimal values very quickly (CPLEX does not go much further down over the next 22 hours). Similarly excellent results are obtained on all other incidents.

\begin{figure}[t]
    \begin{center}
        \epsfig{file=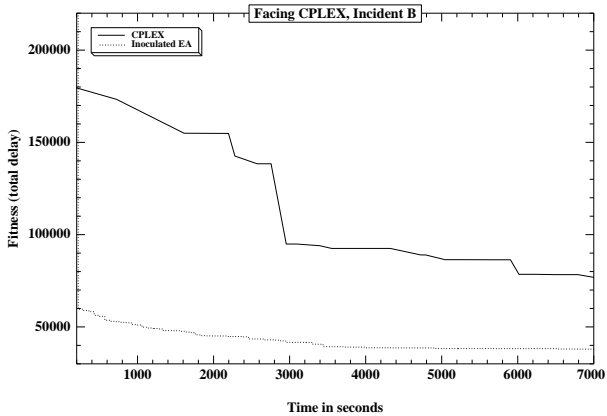,width=8cm}      
      \caption{On this instance, a better quality solution is obtained by the EA after two hours although CPLEX later catches up and goes slightly further down (after 6 hours).}
      \label{fig:facingCPLEX2}
    \end{center}
\vspace{-0.4cm}
\end{figure}

In some cases, over the considered period of two hours, as shown in figure \ref{fig:facingCPLEX2}, the inoculated version of the Evolutionary Algorithm is even able to reach solutions of much better quality than those obtained by CPLEX over the same period of time. In this particular case, CPLEX actually catches up only after 6 hours of computation time and goes only slightly further down. 

\section{Conclusions}
\label{sec:conclusion}
We described a permutation-based evolutionary algorithm that efficiently solves a large and difficult real-world problem. Thanks to a Non-Random Initialization based on the Inoculation of a good starting point obtained after pre-solving part of the problem, the competition with CPLEX is won by far in most cases except when the solution is trivial and found immediately by CPLEX. 

As mentioned earlier, one minute of delay on the railway network costs, everything included, around 1000 euros (\~1200 USD) \cite{RailEtRecherche}. If a situation such as the one illustrated in figure \ref{fig:facingCPLEX1} occurs only once a day, yearly potential savings brought by the Evolutionary Algorithm with respect to CPLEX reach several dozen million dollars.

CPLEX however, remains interesting for two reasons. First it is faster on easy instances. Second, if given enough time, it reaches better quality solution (\cite{semet:CEC05}).  A good practical guideline would therefore be to use two computers and always run the EA and CPLEX in parallel. Moreover, more work on hybridizing both algorithms (see \cite{semet:CEC05} for early efforts in that direction) would certainly be fruitful.

If more computers are available, an additional strength of the Evolutionary Algorithm can be used: parallelization. This can be done in two ways. The first one is to make several runs in parallel. All results considered for this work were averaged out of 11 runs. Inter-run variance being usually quite high, a parallel run on 11 computers would have been very profitable. Parallel runs would, additionally, allow to benefit from several parametrisations (such as $MM$ and $T+R$),thus increasing the chances for a quick convergence. Ultimately, this problem will have to be solved in real-time, which means within a few minutes, almost regardless of the computational ressources needed. The second way to parallelize, which is to distribute fitness evaluations, could therefore be considered. In our case, with a $(10+70)$ replacement scheme, a 70 computers cluster would reduce the considered two hours to less than two minutes.


\section{Acknowledgments}
We gratefully acknowledge funding provided by SNCF's Innovation and Research department under the LIPARI project. Among the SNCF I\&R staff, we would like to personally thank Louis-Marie Cléon, Gilles Dessagne, Christelle L\'erin, Laurent Gely and Véronique de Vulpillières. 
%
{
\bibliographystyle{abbrv}

}
%
%
\end{document}